\documentclass{article}
\usepackage[dvipsnames]{xcolor}
\usepackage{spconf,amsmath,graphicx,hyperref}
\usepackage{amssymb}
\usepackage{tikz}
\usepackage{comment}
\usepackage{multirow}
\usepackage{booktabs}
\usepackage[capitalise]{cleveref} 
\usepackage{pifont}
\newcommand{\cmark}{\ding{51}}%
\newcommand{\xmark}{\ding{55}}%

\def\L{{\cal L}}

\title{DarkVRAI: Capture-Condition Conditioning and Burst-Order Selective Scan for Low-light RAW Video Denoising}
\name{Youngjin Oh \qquad Junhyeong Kwon \qquad Junyoung Park \qquad Nam Ik Cho}
\address{Department of ECE, INMC, Seoul National University, Seoul, Korea}
\begin{document}
%\ninept
%
\maketitle
\begin{abstract}
Low-light RAW video denoising is a fundamentally challenging task due to severe signal degradation caused by high sensor gain and short exposure times, which are inherently limited by video frame rate requirements. To address this, we propose DarkVRAI, a novel framework that achieved first place in the AIM 2025 Low-light RAW Video Denoising Challenge. Our method introduces two primary contributions: (1) a successful application of a conditioning scheme for image denoising, which explicitly leverages capture metadata, to video denoising to guide the alignment and denoising processes, and (2) a Burst-Order Selective Scan (BOSS) mechanism that effectively models long-range temporal dependencies within the noisy video sequence. By synergistically combining these components, DarkVRAI demonstrates state-of-the-art performance on a rigorous and realistic benchmark dataset, setting a new standard for low-light video denoising.
\end{abstract}
\section{Introduction}
\label{sec:intro}

The ability to capture high-quality video in low-light conditions is critical for a wide range of applications, from consumer photography to autonomous systems. However, the physics of image capture imposes significant constraints. Video acquisition requires maintaining a target frame rate (e.g., 24 to 120 fps), which places a strict upper bound on the exposure time for each frame. In dim environments, this short integration time prevents the sensor from gathering a sufficient number of photons, necessitating high electronic gain (ISO) to produce a visible image. This amplification process drastically degrades the signal-to-noise ratio (SNR), as it magnifies not only the weak photon signal but also inherent noise sources like read noise and photon shot noise~\cite{wei2020physics}.

Processing the data in its RAW format, before demosaicing and other in-camera signal processing (ISP) steps, offers the potential for higher-quality restoration~\cite{conde2024toward} but introduces further challenges. RAW image data~\cite{abdelhamed2018high,plotz2017benchmarking} exhibits complex, signal-dependent noise characteristics that are specific to each sensor, and denoising algorithm must operate on the mosaicked Bayer pattern without introducing artifacts. 
The difficulty of restoration is exacerbated in a video scenario as the robust alignment of frames in the presence of scene motion, especially in the extremely low-SNR domain of RAW video, presents a significant problem.

The recent AIM 2025 Low-light RAW Video Denoising Challenge~\cite{aim2025videodenoising} provides a formidable and realistic testbed for evaluating denoising algorithms. Its newly introduced dataset is captured using a precise automated system with 14 distinct smartphone camera sensors under 9 different capture conditions, spanning illuminance levels from 1 to 10 lx and exposure times from 1/120 s to 1/24 s. The use of a diverse set of real-world hardware and conditions, coupled with a rigorous evaluation on a private test set, establishes a challenging new benchmark intended to accelerate future research in low-light RAW video denoising.

In this paper, we provide a detailed description of our proposed framework, DarkVRAI, the winning solution of the AIM 2025 Low-light RAW Video Denoising Challenge. While the method was briefly summarized in the official challenge report~\cite{aim2025videodenoising}, here we offer a comprehensive explanation of its architecture and the design choices that led to its superior performance, and additional in-depth analysis.

Our core contribution lies in successfully extending a conditioning mechanism, previously effective for single-image denoising~\cite{oh2025towards}, to the video domain. We explicitly leverage the `capture conditions'---such as sensor type, scene illuminance, and frame rate---to adaptively guide both the frame alignment and denoising stages. Unlike blind methods that must implicitly infer these properties, our framework uses this prior information to specialize its processing, leading to more effective noise removal across diverse hardware. Furthermore, to enhance temporal aggregation, we introduce the Burst-Order Selective Scan (BOSS) mechanism, which is designed to capture long-range dependencies by scanning frame features in burst order. By combining condition-aware processing with a robust temporal fusion mechanism, DarkVRAI effectively restores high-quality video frames from severely degraded RAW inputs.

We summarize the contributions as follows:
\begin{itemize}
    \item We propose DarkVRAI, a two-stage framework that sets a new state-of-the-art on a novel benchmark for low-light RAW video denoising, validated by its first-place ranking in the AIM 2025 Challenge.
    \item We extend a capture condition guidance mechanism, which injects explicit metadata (sensor, illuminance, frame rate) into the network, for image denoising to a multi-frame task. This enables the model to adapt its behavior to specific degradation characteristics.
    \item We design Burst-Order Selective Scan (BOSS), which leverages principles from state-space models to progressively aggregate long-range temporal information, leading to more robust feature alignment and superior noise suppression.
\end{itemize}

\section{Method}
\label{sec:method}

\begin{figure}[t]
\centering
   \includegraphics[width=7.8cm]{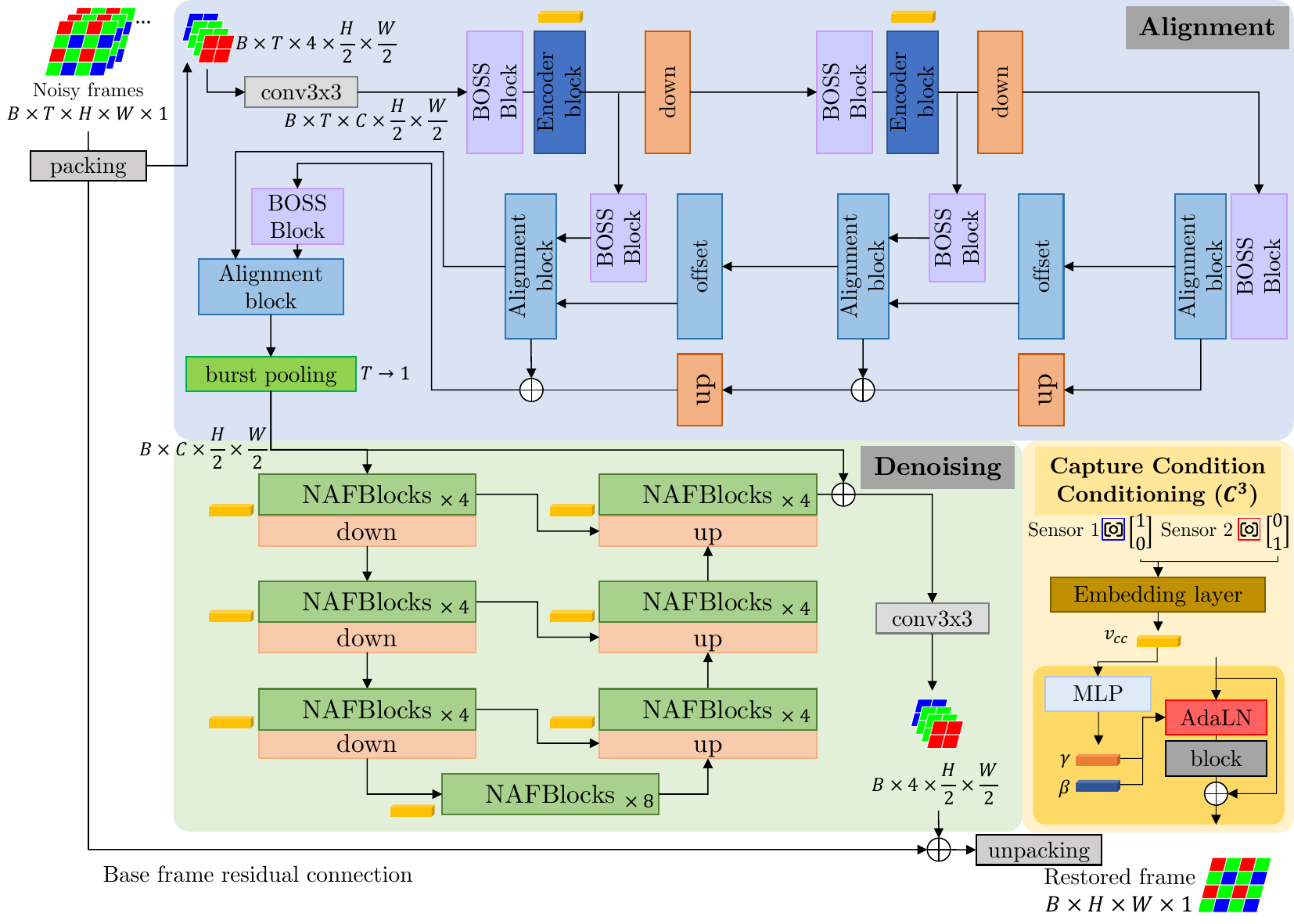}
   \hfil
\caption{Proposed framework for low-light RAW video denoising, DarkVRAI. The framework consists of a frame alignment stage, followed by a denoising stage. Both stages are enhanced with capture condition conditioning (C$^3$).}
\label{figure:figure_framework}
\end{figure}

\begin{figure}[t]
\centering
   \includegraphics[width=7.8cm]{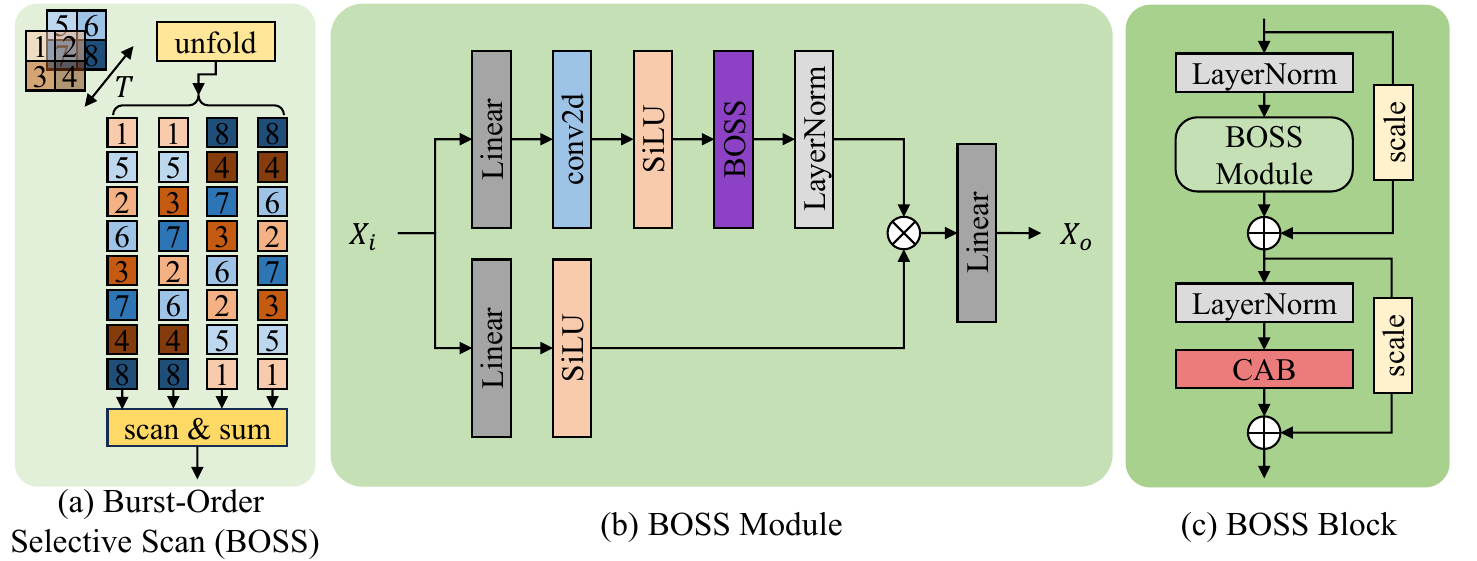}
   \hfil
\caption{Proposed Burst-Order Selective Scan (BOSS), and its hierachical structure: (a) the scanning mechanism, (b) BOSS module, and (c) final BOSS block.}
\label{figure:figure_boss}
\end{figure}

\subsection{Overall Framework}
DarkVRAI is designed as a two-stage pipeline consisting of frame alignment, followed by denoising, as illustrated in \cref{figure:figure_framework}. The framework accepts a sequence of 10 consecutive noisy RAW frames as input and produces a single, clean RAW frame corresponding to the final frame in the sequence. Both stages are modulated by our Capture Condition Conditioning (C$^3$) scheme, which adapts the network's behavior based on explicit metadata about the input video sequence. Additionally, the alignment stage is enhanced with our proposed Burst-Order Selective Scan (BOSS) mechanism for improved temporal context processing. We explain each component in the following subsections.

\subsection{Capture Condition Conditioning}
A key challenge in low-light video denoising is achieving robust performance that generalizes across the vast diversity of modern camera sensors and their capture settings.
A ``blind" denoising model would be suboptimal, as the noise characteristics are not constant; they are a complex function of the sensor's physical properties, specific lighting, and exposure settings~\cite{abdelhamed2018high,plotz2017benchmarking}.
Moreover, video sequences captured with different frame rates would display varying motion between frames according to their frame rate. This variability in motion presents a parallel challenge for the frame alignment stages, which are critical for effective video denoising.

A blind network would be forced to expend a portion of its capacity implicitly inferring these conditions from the noisy input sequences. Our core hypothesis is that by explicitly providing and conditioning the network with information about the degradation context through capture conditions, we enable the model to dedicate its full representational power to frame alignment and denoising themselves.

Inspired by recent work in controllable image denoising~\cite{oh2025towards}, we implement a Capture Condition Conditioning (C$^3$) scheme on video denoising. A simplified procedure is illustrated in \cref{figure:figure_framework}. Each capture condition---sensor type, illuminance (lx), and frame rate (fps)---is first one-hot encoded. These sparse representations are then projected into a dense, trainable degradation embedding layer to generate the capture condition vector $v_{cc}$. This vector is used to modulate the features within both the alignment and denoising networks via adaptive layer normalization (AdaLN)~\cite{peebles2023scalable,ba2016layer}. Given an input and output tensor $x,y\in\mathbb{R}^{B\times C\times H\times W}$, the process is as follows:
\begin{equation}
    \begin{aligned}
    & \mu_b=\frac{1}{CHW}\sum^C_{c=1}\sum^H_{h=1}\sum^W_{w=1}x_{b,c,h,w},\\
    & \sigma_b^2=\frac{1}{CHW}\sum_{c,h,w}(x_{b,c,h,w}-\mu_b)^2,\\
    & y_{b,c,h,w}=\gamma_{b,c}\frac{x_{b,c,h,w}-\mu_b}{\sqrt{\sigma_b^2+\epsilon}}+\beta_{b,c},\\
    \end{aligned}
\end{equation}
where $\mu_b,\sigma_b$ denote the mean and standard variation of the input sample, and $\beta_{b,c},\gamma_{b,c}$ are the channel-wise bias and gain derived from $v_{cc}$.

This process effectively tailors the feature statistics at each layer, allowing the model to dynamically adjust its strategy for varying motion alignment and different noise profiles.

\subsection{Burst-Order Selective Scan}
In low-light video frames, the most powerful tool against noise is the temporal redundancy available across the burst of frames. However, scene motion makes simple averaging or fusion ineffective. To overcome this, an effective model must capture the complex spatio-temporal relationships between frames across the entire sequence. While Transformer-based attention mechanisms~\cite{vaswani2017attention} have proven effective, recent advancements in State-Space Models (SSM)~\cite{gu2023mamba} offer a compelling alternative for modeling long-range dependencies with greater computational efficiency.

Our Burst-Order Selective Scan (BOSS) mechanism adapts this principle for burst video processing. The mechanism, along with the BOSS block and the BOSS module, is shown in \cref{figure:figure_boss}. BOSS scans the sequence of 10 input frame features in their natural temporal order. For an input tensor $X_{i}$, the output of BOSS module $X_{o}$ is:
\begin{equation}
    \begin{aligned}
    & xz=Linear(X_{in})=[x\|z],\\
    & x_{boss}=BOSS(\phi(Conv2d(x))),\\
    & y=LayerNorm(x_{boss}),\\
    & X_o=Linear(y \odot \phi(z)),
    \end{aligned}
\end{equation}
where $\phi$ is the SiLU activation function.

We strategically place the BOSS block, which is made up of our BOSS module and a channel attention block~\cite{woo2018cbam} in sequence, like RSSB~\cite{guo2024mambair}, before each encoder and alignment block within the frame alignment stage. This progressive application is critical, because it ensures that the features being processed within the alignment network are already enriched with long-range temporal context. By understanding the temporal trajectory of features before attempting to align them, the alignment process becomes significantly more robust and accurate, especially in high-noise, low-signal scenarios.

\subsection{Network Architecture}
The architectural backbone of DarkVRAI combines proven components for multi-frame processing and image restoration. The frame alignment stage is built upon the EDA~\cite{dudhane2023burstormer}, which is a strong module for burst image alignment. The subsequent denoising stage employs a U-shaped structure~\cite{ronneberger2015u} built with convolutional NAFBlocks~\cite{chen2022simple}, which are recognized for their high performance and efficiency in restoration. The initial number of channels $C$ is 48, and number of frames $T$ that our framework processes simultaneously is 10. The encoder and decoder stages of the denoising network each contain 4 blocks, while the bottleneck is composed of 8 blocks.

\section{Experiments}
\label{sec:experiments}

\subsection{Experimental Setup}
\noindent\textbf{Dataset.} Experiments are conducted using the AIM 2025 Low-light RAW Video Denoising Challenge dataset~\cite{aim2025videodenoising}. The large-scale dataset consists of 6 unique scenes captured with 14 different smartphone cameras under 9 distinct illumination and exposure conditions, resulting in a total of 756 video sequences. An undisclosed seventh scene is retained for testing.

\noindent\textbf{Evaluation.} Evaluations were performed on CodaBench~\cite{xu2022codabench} with a private validation and test set, ensuring a fair and unbiased comparison of all submitted methods. The metrics are Peak Signal-to-Noise Ratio (PSNR) and Structural Similarity Index Measure (SSIM), computed on the linear mosaicked RAW data to measure the fidelity of the restored output.

\noindent\textbf{Implementation Details.} The model was trained exclusively on the data provided for the AIM 2025 challenge~\cite{aim2025videodenoising}, without using any external datasets. 
No data augmentations such as flipping or rotations are applied in order to preserve the input Bayer pattern. The training data was cropped into patches of size 256$\times$256 with a stride of 192. We trained the model for 300,000 iterations with a batch size of 4 using the Adam optimizer~\cite{kingma2014adam}, with parameters $\beta_1=0.9$ and  $\beta_2=0.999$. The learning rate was initialized at 2e-4 and gradually decayed to 1e-6 using a Cosine Annealing schedule~\cite{loshchilov2016sgdr}. The optimization was guided by a combined loss function of L1 and MS-SSIM~\cite{wang2003multiscale}. All experiments were conducted within the PyTorch framework on NVIDIA RTX 3090 GPUs.

\subsection{Results}
\begin{table}[t]
\centering
\caption{Quantitative results on the AIM 2025 Low-light RAW Video Denoising Challenge private test set. Data is sourced from the official challenge leaderboard.}
\resizebox{0.40\textwidth}{!}{
\begin{tabular}{c|c|cc}
\toprule
Method       & Type                          & PSNR $\uparrow$ & SSIM $\uparrow$ \\ \midrule
Noisy        & -                             & 36.06           & 0.8093          \\ \midrule
UNet~\cite{ronneberger2015u}         & \multirow{3}{*}{Single-Frame} & 43.52           & 0.9691          \\
YOND~\cite{feng2025yond}         &                               & 45.76           & 0.9682          \\
Peng~\cite{chen2022simple}  &                               & 46.52           & 0.9814          \\ \midrule
UNet w/ Attn & \multirow{3}{*}{Multi-Frame}  & 45.72           & 0.9797          \\
Xu \textit{et al.}~\cite{zamir2022restormer}    &                               & 48.19           & 0.9865          \\
DarkVRAI     &                               & \textbf{48.32}  & \textbf{0.9879} \\ 
\bottomrule
\end{tabular}
\label{table:results}
}
\end{table}

\begin{table}[t]
\centering
\caption{Ablation study of C$^3$ and BOSS block on the AIM 2025 Low-light RAW Video Denoising Challenge validation set. \textcolor{Green}{Green text} indicates the gain over the baseline.}
\resizebox{0.48\textwidth}{!}{
\begin{tabular}{c|cc|cc}
\toprule
Model    & C$^3$  & BOSS   & PSNR $\uparrow$ & SSIM $\uparrow$  \\ \midrule
Baseline & \xmark & \xmark & 46.39           & 0.9836           \\
Model A  & \cmark & \xmark & 46.85 \textcolor{Green}{(+0.46)}   & 0.9853  \textcolor{Green}{(+0.0017)} \\
DarkVRAI & \cmark & \cmark & 47.16 \textcolor{Green}{(+0.77)}   & 0.9854  \textcolor{Green}{(+0.0018)} \\ 
\bottomrule
\end{tabular}
\label{table:ablation}
}
\end{table}

\begin{figure*}[t]
\centering
   \includegraphics[width=17.8cm]{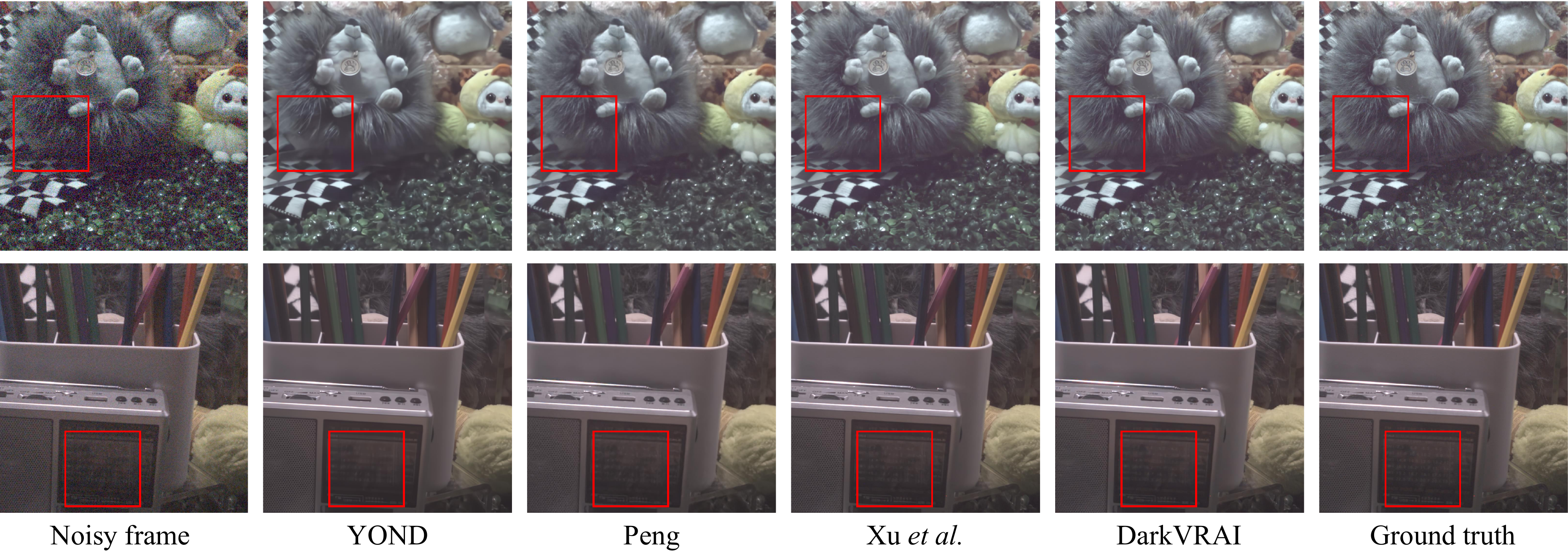}
   \vfil
\caption{Visual comparison of noisy input base frames, and restored outputs of YOND, Peng, Xu \textit{et al.}, DarkVRAI, and the ground truths from the private test set. The results are adapted from the official challenge report~\cite{aim2025videodenoising} with permission. Zoom in for better analysis. Our method produces more detailed frames without excessive denoising compared to competing methods.}
\label{figure:figure_visual}
\end{figure*}

\noindent\textbf{Quantitative Results.} The quantitative results on the private test set are shown in \cref{table:results}. Our method, DarkVRAI, achieved the highest scores in both PSNR and SSIM, securing the first-place ranking in the challenge and demonstrated state-of-the-art performance on the dataset.

A key observation from \cref{table:results} is the substantial performance gap between multi-frame and single-frame approaches. Our method achieves a PSNR of 48.32 dB, which is a significant 1.8 dB higher than the best-performing single-frame method (Peng at 46.52 dB). This tendency is also consistent when comparing the single-frame UNet to its variant UNet w/ Attn, which utilizes channel attention to fuse multiple noisy frames for restoration. This large margin emphasizes the importance of effectively leveraging temporal information for low-light video denoising, which is a core strength of BOSS and C$^3$ strategy.

\noindent\textbf{Qualitative Results.} Visual comparison is available in \cref{figure:figure_visual}. The results are adapted from the official challenge report~\cite{aim2025videodenoising}.
We compare our method with competing single-frame methods YOND~\cite{feng2025yond} and Peng's NAFNet-based~\cite{chen2022simple} model, and Xu \textit{et al.}'s Restormer-based~\cite{zamir2022restormer} multi-frame solution.

The visual results are consistent with the quantitative results. Single-frame approaches (YOND, Peng), while showing significant improvement over the noisy input, exhibits a tendency to over-smooth delicate details.
In contrast, multi-frame methods (Xu \textit{et al.}, DarkVRAI) demonstrates a superior ability to remove heavy noise while simultaneously preserving fine textures and detail.
Furthermore, DarkVRAI produces qualitatively better results to the competing multi-frame method (Xu \textit{et al.}). We attribute this to our framework's advanced temporal aggregation capabilities, compared to a simple input concatenating strategy.

\subsection{Ablation Studies}
To verify the effectiveness of the main components of DarkVRAI, we conducted a series of ablation studies. The results are shown in \cref{table:ablation}. We evaluated the performance of the following three models on the validation set of AIM 2025 challenge dataset; \textit{Baseline}: a model without our two main contributions, \textit{Model A}: the baseline model enhanced with the C$^3$, and \textit{DARKVRAI}: our proposed model, which includes both the C$^3$ and the BOSS mechanism.

By adding the capture condition information, Model A achieved a performance gain of +0.46 dB in PSNR and +0.0017 in SSIM over the Baseline. This demonstrates that explicitly providing the network with sensor and capture settings is beneficial in adapting to the diverse noise characteristics and motion of a video sequence.

Building on this, DarkVRAI incorporates BOSS for improved temporal aggregation. This addition yields a further performance boost, resulting in a total gain of +0.77 dB in PSNR and +0.0018 in SSIM compared to the Baseline. This demonstrates that the BOSS block effectively handles temporal information between frames, which is vital for high-quality video denoising.

\section{Conclusion}
\label{sec:conclusion}
This paper introduced DarkVRAI, our winning solution to the AIM 2025 Low-light RAW Video Denoising Challenge. We have addressed the fundamental difficulties of this task—severe noise, motion, and sensor-specific degradations—through a novel two-stage framework. The key innovations of our method are the synergistic use of the C$^3$ strategy for noise-adaptive processing, and the BOSS for powerful, long-range temporal modeling. The first-place result in a competitive challenge, conducted on a diverse and realistic dataset, validates our design choices and establishes DarkVRAI as a new state-of-the-art method in low-light RAW video denoising.

% -------------------------------------------------------------------------

\end{document}